# Building Chinese Lexicons from Scratch by Unsupervised Short Document Self-Segmentation

Daniel Gayo-Avello
Department of Informatics, University of Oviedo, Calvo Sotelo s/n 33007 Oviedo (SPAIN)
Tel. +34 985 10 50 94 / Fax +34 985 10 33 54
`dani@uniovi.es`

**Abstract.** Chinese text segmentation is a well-known and difficult problem. On one side, there is not a simple notion of "word" in Chinese language making really hard to implement rule-based systems to segment written texts, thus lexicons and statistical information are usually employed to achieve such a task. On the other side, any piece of Chinese text usually includes segments present neither in the lexicons nor in the training data. Even worse, such unseen sequences can be segmented into a number of totally unrelated words making later processing phases difficult. For instance, using a lexicon-based system the sequence 巴罗佐 (*Bāluōzuǒ*, Barroso, current president-designate of the European Commission) can be segmented into 巴 (*bā*, to hope, to wish) and 罗佐 (*luōzuǒ*, an undefined word) changing completely the meaning of the sentence. A new and extremely simple algorithm specially suited to work over short Chinese documents is introduced. This new algorithm performs text "self-segmentation" producing results comparable to those achieved by native speakers without using either lexicons or any statistical information beyond the obtained from the input text. Furthermore, it is really robust for finding new "words", especially proper nouns, and it is well suited to build lexicons from scratch. Some preliminary results are provided in addition to examples of its employment.

## Introduction

Chinese texts are written without any kind of "separator" apart from punctuation marks. This, of course, is not a problem for any human skilled in Chinese reading but remains a challenge for NLP systems. Thus, text segmentation, that is, the process of inserting "spaces" between Chinese "words" is a common previous stage for several tasks, for instance machine translation or information retrieval. However, such segmentation problem is not an easy one since there is not a simple notion of "word" in Chinese language. This makes really hard the development of rule-based systems for Chinese segmentation and therefore two main approaches are commonly followed, namely, lexicon- and statistical-based methods.

Lexicon-based systems rely mainly on pre-built sets of Chinese entries and employ that information along with some heuristics in order to segment the input texts, e.g. Chen and Liu (1992). Main limitation of such systems lies on the fact that a lexicon containing any conceivable Chinese segment is unfeasible since new "words" (proper nouns, brands or neologisms) are continuously coined. On the other side, statistical-based systems "distill" information from training data to help finding the most likely "break points", e.g. Sproat and Shih (1990), Maosong *et al.* (1998) or Fu and Luke (2003). Hybrid approaches combining machine-learning and lexicon knowledge have also been proposed, for instance by Sproat *et al.* (1996)

So, Chinese text segmentation is by no means a closed problem and there is still room for improvement and new proposals. In this paper the so-called self-segmentation algorithm is introduced. This new method can also be considered a hybrid one since it uses both statistical information and lexicon knowledge; however, the lexicon is not built in advance but after processing the input data and the statistical information is not previously obtained by training on external corpora but just from the input text. Because of this, while "classical" Chinese segmenters can work on a few characters the self-segmentation algorithm can only be fully exploited by applying it to short documents (i.e. a few hundred characters). Thus, main application of this new algorithm is not text segmentation but "from-scratch" lexicon building since it performs quite robustly and finds equally easily both common Chinese words and rare entries such as phonetic spellings of non-Chinese proper nouns.

## The Chinese Self-Segmentation Algorithm

As it has been said, the proposed algorithm does not employ external lexicons or corpus information beyond the obtained from the input data. The operation is quite simple. First, raw input text must be split into a sequence of chunks using as break points roman characters, numbers and punctuation marks (see Fig. 1).

<<PUT FIGURE 1 HERE>>

**Fig. 1.** Chinese text must be split before being submitted to self-segmentation algorithm.

This preprocessed text is submitted to the self-segmentation algorithm which processes it in sequential stages. First, it finds every *n*-gram (n>2) appearing in the text twice or more times storing both the *n*-gram and its count. Secondly, for each chunk from the input data it finds every *n*-gram appearing within the chunk, the so-called "candidate" *n*-grams. Once all the candidates for a chunk have been found they are "tested" in decreasing length and appearance order (e.g. 巴罗佐 is checked before 巴罗). The test for each *n*-gram is very simple, if counts for both constituent *(n-1)*-grams are strictly bigger than the *n*-gram count then such *n*-gram candidate is "unacceptable" otherwise the *(n-1)*-grams are the "unacceptable" ones. When all the candidates have been tested and a list of acceptable *n*-grams has been found the algorithm goes through that list in decreasing length and count order introducing spaces before and after each acceptable *n*-gram found within the chunk (see Fig. 2)

<<PUT FIGURE 2 HERE>>

**Fig. 2.** A piece of Chinese text is "self-segmented" using information from its original document. 斯特劳 (*Sītèláo*) does not appear in the CEDICT dictionary, it is a phonetic spelling of the proper noun Straw (current UK Foreign Secretary).

During this later stage every segment found within the chunks is added to a "pre-lexicon" if the segment appears twice or more times in the whole segmented text. This pre-lexicon is used afterwards to segment again each chunk. This process is analogous to the one applied with acceptable *n*-grams, that is, traversing the lexicon in decreasing length and count order. This new segmentation phase can also introduce new entries to the lexicon. Thus, the final output of this algorithm includes both a segmented Chinese document and a tentative lexicon. Some preliminary results are provided below. It must be said, however, that by its own nature this algorithm does not always provide valid segments according to Wu's Monotonicity Principle (Wu, 1998) and that the built lexicons, although quite accurate, usually include not only "words" but also complex sentences. A way to overcome these problems is discussed later while the self-segmentation algorithm is provided in the Appendix.

## Preliminary results

Thorough testing of this algorithm over standardized corpora is needed but preliminary results are highly encouraging. Applied over five web pages[1] from the Chinese version of *BBC News* website it has been found that (in average): (1) around 70% of the entries within the tentative lexicons appear in the CEDICT[2] lexicon, (2) around 83% of the lexicon entries actually appearing within the news items also appear in the CEDICT, and (3) at least 37% of the lexicon entries from the news items are correctly segmented since their translations occur literally in the English versions of the original web pages. Table 1 provides more details about these claims while Table 2 shows some entries from the tentative lexicons built for those short documents and Figure 3 includes some selected sentences (headlines, leads and captions) comparing their self-segmentation with results produced by native speakers.

**Table 1.** Whole text from five web pages was self-segmented. These are the number of entries from the web pages that appear in CEDICT dictionary, the number of entries which are actually from the news items and appear in CEDICT and the number of literal translations of Chinese segments into English words that appear in the English version of the same news items.

<<PUT TABLE 1 HERE>>

---

[1] Arafat in France: `http://news.bbc.co.uk/chinese/simp/hi/newsid_3960000/newsid_3963900/3963989.stm`
British Gov. and Linux: `http://news.bbc.co.uk/chinese/simp/hi/newsid_3960000/newsid_3961800/3961891.stm`
Bush Website: `http://news.bbc.co.uk/chinese/simp/hi/newsid_3960000/newsid_3963800/3963879.stm`
Buttiglione: `http://news.bbc.co.uk/chinese/simp/hi/newsid_3950000/newsid_3959100/3959187.stm`
EU New Constitution: `http://news.bbc.co.uk/chinese/simp/hi/newsid_3960000/newsid_3964000/3964071.stm`

[2] CEDICT is a public-domain Chinese-English dictionary started by Paul Denisowski and now available at `http://www.mandarintools.com/cedict.html`

**Table 2.** Tentative lexicons built by self-segmenting previously mentioned web pages. Only entries appearing twice or more times in the news items and with their corresponding translations occurring in English version are shown. Entries which are not available in CEDICT are shown in bold.

<<PUT TABLE 2 HERE>>

<<PUT FIGURE 3 HERE>>

**Fig. 3.** Some headlines, leads and captions self-segmented within previous web pages. The algorithm results are compared with human-made segmentations. Matches between both ones are noteworthy, specially taking into account that the text used by the algorithm is so scarce (around 800 characters each document).

## Discussion

As it can be seen from above preliminary results the segmentation achieved by this new algorithm is similar to that produced by human beings. In contrast to state-of-the-art segmenters, which rely on lexicons and statistical information extracted from external corpora, the proposed algorithm simply uses measures obtained from the input text and not only does not employ external lexicons but can be used to build Chinese lexicons from scratch. Nevertheless, by using short Chinese documents some sentences are not totally segmented and thus some entries from the built lexicons are not "words" but complex statements.

In order to apply this algorithm to develop an accurate Chinese lexicon such issues can be overcome by finding (e.g. using search engines) unseen documents containing one or more "promising"[3] entries from the tentative lexicon. This new documents can be self-segmented obtaining reinforcement for some entries within the lexicon in addition to new entries. Table 3 shows some reinforced and new lexicon entries obtained from a USENET post reached by using the keyword 阿拉法特 (*Ālāfǎtè*, Arafat).

**Table 3.** A new lexicon built from a post published in `talk.politics.china` on May 28$^{th}$ 2002. Only entries appearing twice or more times in the post are shown. New entries to CEDICT are shown in bold while "reinforced" entries (i.e. entries appearing both in original and new document) appear with grey background.

<<PUT TABLE 3 HERE>>

## Conclusion

An extremely simple algorithm to perform self-segmentation of Chinese documents has been described. This algorithm does not use either external lexicons or statistical information apart from the input data. Nevertheless, it provides segmentation results comparable to those achieved by native speakers and, in addition to this, rather accurate lexicon entries. Such algorithm can be applied to build Chinese lexicons from scratch beginning with some seed documents and relying on search engines to find new unseen but related documents which will provide reinforcement for some of the entries besides increasing the lexicon size with new entries. The algorithm could also be applied by other segmentation techniques as a preprocessing stage since it is quite robust when facing proper nouns.

## Acknowledgments

The author wishes to thank Luo Li and Ho-Meng Chang Shu by their help with the original Chinese texts.

## References

Chen, K.J, and Liu, S.H. (1992) Word Identification for Mandarin Chinese Sentences. In Proc. of COLING-92, pp. 101-107.

Fu, G., and Luke, K.K. (2003) A two-stage statistical word segmentation system for Chinese. Proceedings of the Second SIGHAN Workshop on Chinese Language Processing, pp. 156-159.

Maosong, S., Dayang, S., and Tsou, B.K. (1998) Chinese Word Segmentation without Using Lexicon and Hand-crafted Training Data. Proceedings of the 36$^{th}$ Annual Meeting of the ACL, pp. 1265-1271.

Sproat, R., and Shih, C. (1990) A statistical method for finding word boundaries in Chinese text. Computer Processing of Chinese and Oriental Languages, 4, pp. 336-351.

---

[3] The longest and the most frequent the entry the most promising it is.

Sproat, R., Shih, C., Gale, W., and Chang, N. (1996) A Stochastic Finite-State Word-Segmentation Algorithm for Chinese. Computational Linguistics, vol. 22, no. 3, pp. 377-404.

Wu, D. (1998) A Position Statement on Chinese Segmentation. 1st Chinese Language Processing Workshop, University of Pennsylvania, Philadelphia. http://www.cs.ust.hk/~dekai/papers/segmentation.html

# Appendix

*Algorithm selfSegmentation (chineseChunks)*
**Input:** Chinese text split into the list *chineseChunks* using roman characters, numbers and punctuation marks as break-points

1. **from** *n* ← *2* **to** *MAX_NGRAM_SIZE* **do**
2.    **for each** *n-gram ngram found in chineseChunks* **do**
3.       *weights(ngram)* ← **count***(chineseChunks, ngram)*
4.    **loop**
5. **loop**
6. **for each** *chunk chunk in chineseChunks* **do**
7.    **if** *length of chunk > 2*
8.       **from** *n* ← *2* **to** *length of chunk* **do**
9.          **for each** *n-gram ngram found in chunk* **do**
10.             *count* ← *weights(ngram)*
11.             **if** *count > 1*
12.                *candidates(n)(ngram)* ← *count*
13.             **end if**
14.          **loop**
15.       **loop**
16.       **if** *there exist candidates*
17.          **from** *n* ← *size of candidates+1* **to** *2* **do**
18.             **for each** *candidate ncandidate in candidates(n)* **do**
19.                **for each** *candidate n1candidate in candidates(n-1)* **do**
20.                   **if** *n1candidate found in ncandidate*
21.                      **if** *weights(n1candidate) > weights(ncandidate)*
22.                         *ncandidate is unacceptable*
23.                      **else**
24.                         *n1candidate is unacceptable*
25.                      **end if**
26.                 **end if**
27.                **loop**
28.             **loop**
29.          **loop**
30.          **from** *n* ← *size of candidates+1* **to** *2* **do**
31.             **for each** *candidate ncandidate in candidates(n)* **do**
32.                **if** *ncandidate is not unacceptable*
33.                   **sortedInsert** *(acceptables, ncandidate)*
34.                **end if**
35.             **loop**
36.          **loop**
37.          *segmentedChunk* ← *chunk*
38.          **for each** *candidate candidate in acceptables* **do**
39.             *segmentedChunk* ← *replace(candidate, <BLANK>candidate<BLANK>, segmentedChunk)*
40.          **loop**
41.          *chunk* ← *segmentedChunk*
42.          *lexiconEntries* ← *explode(<BLANK>, chunk)*
43.          **for each** *entry entry in lexiconEntries* **do**
44.             *lexicon(entry)* ← *lexicon(entry)+1*
45.          **loop**
46.    **end if**
47. **end if**

48. **loop**
49. **for each** entry entry in lexicon **do**
50.     **if** lexicon(entry) = 1
51.         drop lexicon(entry)
52.     **end if**
53. **loop**
54. sort lexicon by descending length and count order
55. **for each** chunk chunk in chineseChunks **do**
56.     segmentedChunk ← chunk
57.     **for each** entry entry in lexicon **do**
58.         segmentedChunk ← replace(entry, <BLANK>entry<BLANK>, segmentedChunk)
59.     **loop**
60.     chunk ← segmentedChunk
61.     lexiconEntries ← explode(<BLANK>, chunk)
62.     **for each** entry entry in lexiconEntries **do**
63.         lexicon(entry) ← lexicon(entry)+1
64.     **loop**
65. **loop**
66. **for each** entry entry in lexicon **do**
67.     **if** lexicon(entry) = 1
68.         drop lexicon(entry)
69.     **end if**
70. **loop**
71. sort lexicon by descending length and count order
72. **return** lexicon

**Input data:** 英国外相斯特劳对BBC说，英国可能将会在2006年举行公民投票。
**Split data:**
1. 英国外相斯特劳对
2. 说
3. 英国可能将会在
4. 年举行公民投票

---

**Input data:** 英国外相斯特劳对

**Candidate n-grams:**

| n-gram | Weight in whole document |
|---|---|
| 斯特劳 | 2 |
| 英国 | 3 |
| 斯特 | 2 |
| 特劳 | 2 |

**Is 斯特劳 acceptable?**

| | n-gram | 斯特劳 | weight | 2 |
|---|---|---|---|---|
| 1st constituent | 斯特 | weight | 2 |
| 2nd constituent | 特劳 | weight | 2 |

Yes, 斯特劳 is acceptable

**Acceptable n-grams:**
斯特劳   2
英国     3

**Segmentation steps:**
0. 英国外相斯特劳对
1. 英国外相 · 斯特劳 · 对
2. 英国 · 外相 · 斯特劳 · 对

**Output:** 英国 · 外相 · 斯特劳 · 对

**Naïve literal translation:** English · Foreign Minister · *Straw* · to

| News item | Total entries found in CEDICT | Entries from news items found in CEDICT | Translations of segments found in English version |
|---|---|---|---|
| Arafat in France | 54/77 **(70%)** | 40/48 **(83%)** | 15/48 **(31%)** |
| British Government Linux | 45/64 **(70%)** | 34/37 **(92%)** | 17/37 **(46%)** |
| Bush Website | 42/63 **(67%)** | 28/33 **(85%)** | 14/33 **(42%)** |
| Buttiglione | 81/114 **(71%)** | 67/85 **(79%)** | 21/85 **(25%)** |
| EU New Constitution | 57/82 **(70%)** | 40/49 **(82%)** | 17/49 **(35%)** |
| **Average** | **279/400 (70%)** | **209/252 (83%)** | **98/252 (37%)** |

| | Arafat Emergency Treatment France | | |
|---|---|---|---|
| 阿拉法特 | Ālāfǎtè | Arafat (Yasser) | |
| 巴勒斯坦 | Bālèsītǎn | Palestine | |
| 拉姆安拉 | Lāmǔ'ānlā | Ramallah | |
| 以色列 | Yǐsèliè | Israel | |
| 法国 | Fǎguó | France / French | |
| 治疗 | zhìliáo | to treat / to cure / (medical) treatement / cure | |
| 医院 | yīyuàn | hospital | |
| 医生 | yīshēng | doctor | |
| 星期 | xīngqī | week | |
| 巴黎 | Bālí | Paris | |
| 血液 | xuèyè | blood | |
| 否认 | fǒurèn | to declare to be untrue / to deny | |
| 紧急 | jǐnjí | urgent | |
| 总理 | zǒnglǐ | premier / prime minister | |
| 领导 | lǐngdǎo | lead / leading / to lead / leadership | |

| | British Government and Linux | | |
|---|---|---|---|
| 政府 | zhèngfǔ | government | |
| 英国 | Yīngguó | England / Britain / English | |
| 软件 | ruǎnjiàn | (computer) software | |
| 使用 | shǐyòng | to use / to employ / to apply / to make use of | |
| 微软 | Wēiruǎn | Microsoft | |
| 中国 | Zhōngguó | China / Chinese | |
| 公司 | gōngsī | (business) company / company / firm / corporation / incorporated | |
| 产品 | chǎnpǐn | goods / merchandise / product | |
| 机构 | jīgòu | organization / agency / institution | |
| 减少 | jiǎnshǎo | to lessen / to decrease / to reduce / to lower | |
| 系统 | xìtǒng | System | |
| 视窗 | Shìchuāng | Windows (operating system) | |
| 免费 | miǎnfèi | free (of charge) | |
| 国家 | guójiā | country / nation | |
| 安装 | ānzhuāng | install / erect / fix / mount / installation | |
| 开发 | kāifā | exploit (a resource) / open up (for development) / to develop | |
| 依赖 | yīlài | to depend on / to be dependent on | |
| 国 | guó | country / state / nation | |

| | EU New Constitution | | |
|---|---|---|---|
| 成员国 | chéngyuán-guó | member country | |
| 斯特劳 | Sītèláo | Straw (Jack) | |
| 欧盟 | Ōuméng | European Union | |
| 英国 | Yīngguó | England / Britain / English | |
| 宪法 | xiànfǎ | constitution (of a country) | |
| 成员 | chéngyuán | member | |
| 欧洲 | Ōuzhōu | Europe / Europen | |
| 签署 | qiānshǔ | to sign (an agreement) | |
| 罗马 | Luómǎ | Rome / Roman | |
| 领袖 | lǐngxiù | leader | |
| 条约 | tiáoyuē | treaty / pact | |
| 议会 | yìhuì | parliament / legislative assembly | |
| 主席 | zhǔxí | chairperson / premier / chairman | |
| 举行 | jǔxíng | to hold (a meeting, ceremony, etc.) | |
| 联盟 | liánméng | alliance / union / coalition | |
| 仪式 | yíshì | ceremony | |
| 国 | guó | country / state / nation | |
| 新 | xīn | new / newly | |

| | Barroso Buttiglione Problem | | |
|---|---|---|---|
| 巴提格里欧尼 | Bātígélǐ'ōu'ní | Buttiglione (Rocco) | |
| 巴罗佐 | Bāluōzuǒ | Barroso (Jose Manuel) | |
| 意大利 | Yìdàlì | Italy / Italian | |
| 欧盟 | Ōuméng | European Union | |
| 议会 | yìhuì | parliament / legislative assembly | |
| 欧洲 | Ōuzhōu | Europe / Europen | |
| 议员 | yìyuán | member (of a legislative body) / representative | |
| 提名 | tímíng | nominate | |
| 主席 | zhǔxí | chairperson / premier / chairman | |
| 名单 | míngdān | list (of names) | |
| 撤回 | chèhuí | withdraw | |
| 否决 | fǒujué | veto | |
| 接受 | jiēshòu | to accept / to receive | |
| 人选 | rénxuǎn | person(s) selected (for a job, etc.) | |
| 投票 | tóupiào | to vote / vote | |
| 政治 | zhèngzhì | politics / political | |
| 危机 | wēijī | crisis | |
| 部长 | bùzhǎng | head of a (government, etc) department / section chief / section head / secretary / minister | |
| 退出 | tuìchū | to withdraw / to abort / to quit | |
| 争议 | zhēngyì | controversy / dispute | |
| 同意 | tóngyì | to agree / to consent / to approve | |
| 新 | xīn | new / newly | |

| | Bush Campaign Website Blocked | | |
|---|---|---|---|
| 发言人 | fāyánrén | spokes person | |
| 竞选 | jìngxuǎn | run for (electoral) office / campaign | |
| 布什 | Bùshí | Bush (George) | |
| 网站 | wǎngzhàn | network station / node / website | |
| 境外 | jìngwài | outside (a country's) borders | |
| 封锁 | fēngsuǒ | blockade / seal off | |
| 美国 | Měiguó | America / American / United States / USA | |
| 官方 | guānfāng | official / (by the) government | |
| 总统 | zǒngtǒng | president (of a country) | |
| 原因 | yuányīn | cause / origin / root cause / reason | |
| 发表 | fābiǎo | to issue (a statement) / to publish / to issue / to put out | |
| 安全 | ānquán | safe / secure / safety / security | |
| 网络 | wǎngluò | (computer, telecom, etc.) network | |
| 攻击 | gōngjī | to attack / to accuse / to charge / (military) attack | |
| 导致 | dǎozhì | to lead to / to create / to cause / to bring about | |
| 许可 | xǔkě | to allow / to permit | |
| 透露 | tòulù | to leak out / to divulge / to reveal | |

| Original | 阿拉法特在法国接受紧急治疗 |
|---|---|
| Pinyin | Alāfǎtè zài Fǎguó jiēshòu jǐnjí zhìliáo. |
| English | Arafat receives urgent medical treatment in France. |
| Native seg. | 阿拉法特 / 在 / 法国 / 接受 / 紧急 / 治疗 |
| **Self-seg.** | 阿拉法特 / 在 / 法国 / 接受 / 紧急 / 治疗 |
| Original | 阿拉法特乘坐直升机离开拉姆安拉，再转往法国就医 |
| Pinyin | Alāfǎtè chéngzuò zhíshēngjī líkāi Lāmǔānlā, zài zhuǎnwǎng Fǎguó jiùyī |
| English | Arafat departs Ramallah by helicopter, and then goes to France to get treated. |
| Native seg. | 阿拉法特 / 乘坐 / **直升机** / 离开 / 拉姆安拉 , / **再 / 转往** / 法国 / 就医 |
| **Self-seg.** | 阿拉法特 / 乘坐 / **直升 / 机** / 离开 / 拉姆安拉 / **再转往** / 法国 / 就医 |
| Original | 美国总统布什竞选班子发言人透露，布什的官方竞选网站是出于"安全"原因才对境外访问者关闭。 |
| Pinyin | Měiguó zǒngtǒng Bùshí jìngxuǎn bānzi fāyánrén tòulù, Bùshíde guānfān jìngxuǎn wǎngzhàn shì chūyú ānquán yuányīn cái duì jìngwài fǎngwènzhě guānbì. |
| English | The spokesperson for the campaign team of US President Bush disclosed that Bush's official campaign website was blocked to overseas visitors due to "security" reasons. |
| Native seg. | 美国 / 总统 / 布什 / 竞选 / 班子 / 发言人 / 透露 **布什的** / 官方 / 竞选 / 网站 **是 / 出于** / "安全" / 原因 / 才 / 对 / 境外 / **访问者 / 关闭** 。 |
| **Self-seg.** | 美国 / 总统 / 布什 / 竞选 / 班子 / 发言人 / 透露 **布什 / 的** / 官方 / 竞选 / 网站 / **是出于** / 安全 / 原因 / 才 / 对 / 境外 / **访问者关闭** |
| Original | 境外人看到的是"你没有获得进入该网页许可" |
| Pinyin | Jìngwài rén kàndàode shì "Nǐ méiyǒu huòdé jìnrù gāi wǎngyè xǔkě" |
| English | What overseas visitors saw is "You are not authorized to view this page". |
| Native seg. | 境外 / 人 / **看到的** / 是 / **"你 / 没有 / 获得** / 进入 / 该 / 网页 / 许可" |
| **Self-seg.** | 境外 / 人 / **看到 / 的** / 是 / **你没有获得** / 进入 / 该 / 网页 / 许可 |
| Original | 欧盟委员会候任主席巴罗佐撤回了欧盟委员会专员名单提名，巴罗佐的决定是在欧洲议会今天即将对此进行表决前夕作出的。 |
| Pinyin | Ōuméng Wěiyuánhuì hòurèn zhǔxí Bāluōzuǒ chèhuíle Ōuméng wěiyuánhuìzhuānyuán míngdān tímíng, Bāluōzuǒ juédìng shì zài Ōuzhōu Yìhuì jīntiān jíjiāng duìcǐ jìnxíng biǎojué qiánxī zuòchūde. |
| English | The president designate of the European Commission Barroso withdrew the list of nominated EU commissioners, Barroso's decision was taken on the eve before the voting about this issue to be held today in the European Parliament. |
| Native seg. | 欧盟 / **委员会** / 候任 / 主席 / 巴罗佐 / **撤回了** / 欧盟 / **委员会专员** / 名单 / 提名 , **巴罗佐的** / 决定 / 是 / 在 / 欧洲 / **议会** / **今天 / 即将 / 对此** / 进行 / **表决 / 前夕 / 作出的** 。 |
| **Self-seg.** | 欧盟 / **委员** / **会** / 候任 / 主席 / 巴罗佐 / **撤回** / 了 / 欧盟 / **委员** / **会** / **专员** / 名单 / 提名 / **巴罗佐** / **的** / 决定 / 是 / 在 / 欧洲 / **议** / **会** / **今天即将对此** / 进行 / **表决前夕作出** / **的** |
| Original | 巴提格里欧尼拒绝退出提名 |
| Pinyin | Bātígélǐ'ōu'ní jùjué tuìchū tímíng. |
| English | Buttiglione refused to quit from the nomination |
| Native seg. | 巴提格里欧尼 / 拒绝 / 退出 / 提名 |
| **Self-seg.** | 巴提格里欧尼 / 拒绝 / 退出 / 提名 |
| Original | 欧盟25国领袖罗马签署新宪法 |
| Pinyin | Ōuméng 25 guó lǐngxiù Luómǎ qiānshǔ xīn Xiànfǎ |
| English | Leaders of the 25 EU countries signed the new Constitution in Rome |
| Native seg. | 欧盟 / 25 / 国 / 领袖 / 罗马 / 签署 / 新 / 宪法 |
| **Self-seg.** | 欧盟 / 25 / 国 / 领袖 / 罗马 / 签署 / 新 / 宪法 |
| Original | 欧洲联盟25国领袖已经在星期五（29日）早上在罗马签署了宪法条约。 |
| Pinyin | Ōuzhōu Liánméng 25 guó lǐngxiù yǐjīng zài xīngqīwǔ (29 rì) zǎoshang zài Luómǎ qiānshǔle xiànfǎ tiáoyuē. |
| English | Leaders of the 25 countries of the European Union signed the constitutional treaty on friday (29th) morning in Rome. |
| Native seg. | 欧洲 / 联盟 / 25 / 国 / 领袖 / **已经** / **在** / **星期五** / (29日) **早上 / 在** / 罗马 / **签署了** / 宪法 / 条约 。 |
| **Self-seg.** | 欧洲 / 联盟 / 25 / 国 / 领袖 / **已经在** / **星期** / **五** / 29 / 日 / **早上在** / 罗马 / **签署** / **了** / 宪法 / 条约 |
| Original | 英政府机构可望使用LINUX软件 |
| Pinyin | Yīng zhèngfǔ jīgòu kěwàng shǐyòng LINUX ruǎnjiàn |
| English | LINUX software is expected bo be used by British government institutions |
| Native seg. | 英 / 政府 / 机构 / 可望 / 使用 / LINUX / 软件 |
| **Self-seg.** | 英 / 政府 / 机构 / 可望 / 使用 / LINUX / 软件 |
| Original | 英国政府商业办公室一份报告认为：使用LINUX产品可以大幅减少开支，因此英国政府机构使用LINUX产品具有"可行性"。 |
| Pinyin | Yīngguó Zhèngfǔ Shāngyè Bàngōngshì yīfèn bàogào rènwéi: shǐyòng LINUX chǎnpǐn kěyǐ dàfú jiǎnshǎo kāizhī, yīncǐ Yīngguó zhèngfǔ jīgòu shǐyòng LINUX chǎnpǐn jùyǒu "kěxíngxìng". |
| English | A report from the British Office of Government Commerce considers that: using LINUX products can largely reduce costs, therefore it's "feasible" the use of LINUX products by British Government institutions. |
| Native seg. | 英国 / 政府 / **商业** / **办公室** / **一份** / **报告** / 认为：使用 / LINUX / 产品 / 可以 / 大幅 / 减少 / 开支 , 因此 / 英国 / 政府 / 机构 / 使用 / LINUX / 产品 / 具有 / "可行性"。 |
| **Self-seg.** | 英 / **国** / 政府 / **商业办公室一** / **份报告** / 认为 / 使用 / LINUX / 产品 / 可以 / 大幅 / 减少 / 开支 / 因此 / 英 / **国** / 政府 / 机构 / 使用 / LINUX / 产品 / 具有 / 可行性 |

| 中国网民给阿拉法特先生的一封信 |||
| Zhōngguó wǎngmín gěi Ālāfǎtè Xiānsheng de yīfēng xìn |||
| A letter from Chinese internauts to Mr. Arafat |||
|---|---|---|
| 巴勒斯坦 | Bālèsītǎn | Palestine |
| 耶路撒冷 | Yēlùsālěng | Jerusalem |
| 前所未有 | qiánsuǒwèiyǒu | unprecedented |
| 以色列 | Yǐsèliè | Israel |
| **巴拉克** | **Bālākè** | **Barak (Ehud)** |
| 领导人 | lǐngdǎorén | leader |
| 穆斯林 | mùsīlín | Muslim |
| **戴维营** | **Dàiwéiyíng** | **Camp David** |
| **伊斯兰** | **Yīsīlán** | **Islam** |
| 和平 | hépíng | peace |
| 先生 | xiānsheng | sir / mister / teacher / (title of respect) |
| 冲突 | chōngtū | conflict / clash of opposing forces / contention |
| 让步 | ràngbù | (make a) concession / to give in / to yield |
| 主席 | zhǔxí | chairperson / premier / chairman |
| **犹太** | **Yóutài** | **Judea** |
| **拉宾** | **Lābīn** | **Rabin (Isaac)** |
| 民族 | mínzú | nationality / ethnic group |
| 总理 | zǒnglǐ | premier / prime minister |
| 暴力 | bàolì | violence / (use) force / violent |
| 圣殿 | shèngdiàn | temple |
| 协议 | xiéyì | agreement / pact / protocol |
| 面对 | miànduì | confront / face |
| 诚意 | chéngyì | sincerity / good faith |
| **沙龙** | **Shālóng** | **Sharon (Ariel)** |
| 谈判 | tánpàn | to negotiate / negotiation / conference |
| 武装 | wǔzhuāng | arms / equipment / to arm / military / armed (forces) |
| 领导 | lǐngdǎo | lead / leading / to lead / leadership |
| 违约 | wéiyuē | to break a promise / to violate an agreement |
| 中东 | Zhōngdōng | Middle East |
| 签字 | qiānzì | to sign (a signature) |
| 武器 | wǔqì | weapon / arms |